\documentclass{article}  
\usepackage[numbers, compress]{natbib}  
\usepackage[final]{neurips_2019}  
\usepackage[utf8]{inputenc}  
\usepackage[T1]{fontenc}  
\usepackage{hyperref}  
\usepackage{url}  
\usepackage{booktabs}  
\usepackage{amsfonts}  
\usepackage{amsmath}  
\usepackage{nicefrac} 
\usepackage{microtype} 
\usepackage{graphicx}  
\usepackage{subcaption}
\usepackage[inline, shortlabels]{enumitem} 
\newcommand*{\affmark}[1][*]{\textsuperscript{#1}} 

\title{FastEstimator: A Deep Learning Library for Fast Prototyping and Productization}  

\author{ Xiaomeng Dong,~ Junpyo Hong,~ Hsi-Ming Chang,~ Michael Potter,~ Aritra Chowdhury\affmark[*],~ \And Purujit Bahl,~ Vivek Soni,~ Yun-Chan Tsai,~ Rajesh Tamada,~ Gaurav Kumar,~ \And Caroline Favart,~ V. Ratna Saripalli,~ Gopal Avinash\\ GE Healthcare \\ \affmark[*]GE Research} 

\begin{document}  

\maketitle  

\begin{abstract} 
As the complexity of state-of-the-art deep learning models increases by the month, implementation, interpretation, and traceability become ever-more-burdensome challenges for AI practitioners around the world. Several AI frameworks have risen in an effort to stem this tide, but the steady advance of the field has begun to test the bounds of their flexibility, expressiveness, and ease of use. To address these concerns, we introduce a radically flexible high-level open source deep learning framework for both research and industry. We introduce FastEstimator.
\end{abstract} 

\section{Introduction} \label{sec:intro}  
Over the last two decades, interest in Artificial Intelligence (AI) has grown significantly~\cite{ai-report}. This rise in interest has greatly expanded the AI community and created an ever increasing demand for AI systems. Simultaneously, the requirements of these systems are becoming diversified, which poses a challenge in designing a general framework for the AI community.  

Researchers and experts require ultimate flexibility from AI systems since their goal is to explore new ideas and discover uncharted territory in AI. Frameworks like TensorFlow~\cite{tensorflow}, Caffe~\cite{caffe}, MXNet~\cite{mxnet}, CNTK~\cite{cntk}, and PyTorch~\cite{pytorch} gained the favor of expert users because they make few assumptions regarding user behaviors and allow users to control fine-grained details by building experiments from the ground up. However, building AI from scratch may result in unnecessary verbosity and redundant efforts. Therefore, a framework which preserves flexibility while removing these redundancies will be preferable for experts and researchers\textemdash improving their productivity.  

Entrepreneurs, beginners, and enterprise users tend to favor AI systems that have lower learning curves and faster time to deployment. High-level frameworks like Keras~\cite{keras}, Gluon~\cite{mxnet}, fastai~\cite{fastai}, and Ludwig~\cite{ludwig} are examples of such systems. The benefit of a higher-level framework is obviously ease of use; however, simplicity comes at the expense of flexibility. Furthermore, as more and more new ideas in AI have proven useful in real-world applications, high-level frameworks are not evolving fast enough to serve the industry’s interests in these ideas. For example, there are few high-level frameworks that provide flexible support for generative adversarial network (GAN) applications and progressive training schemes. As a result, gaps are forming between state-of-the-art (SOTA) and ease of use.  

We therefore introduce FastEstimator, an open source high-level deep learning library made for both research and industry. Our intent is to bridge the gap by providing a simple yet flexible interface for implementing SOTA ideas, which we hope will help different groups within the AI community to move forward together. For researchers, FastEstimator will continuously monitor the latest advancements in AI to provide an easy and flexible interface. For industry, FastEstimator can enable more SOTA AI products and shorten the product development cycle.  

\section{Highlights of FastEstimator} \label{sec:highlight}  
\textbf{Multi-Framework.} FastEstimator is built on TensorFlow2~\cite{tensorflow} and PyTorch~\cite{pytorch} with a unique multi-framework design. Libraries like Keras~\cite{keras} handle multiple frameworks by unifying different backends with consistent APIs. In contrast, FastEstimator uses the best components from different backends. For example, our data preprocessing module uses a combination of \texttt{torch.DataLoader} and \texttt{tf.data}. As a result, FastEstimator can leverage the advantages of both frameworks and provide a good trade-off between speed and flexibility.

\textbf{Simple yet Flexible.} FastEstimator users only need to interact with three main APIs (see Sec.~\ref{sec:overview}) for any deep learning task. Besides these, the \texttt{Operator} module (see Sec.~\ref{subsec:operator}) allows users to define a complex computational graph in a concise manner. The \texttt{Trace} module (see Sec.~\ref{subsec:trace}) provides users with further control over the training loop. We will show in Sec.~\ref{sec:examples} that FastEstimator can significantly reduce the effort required to implement several deep learning tasks.  

\textbf{Easy to Scale.} Distributed training in many deep learning frameworks requires non-trivial effort on the user’s side. For example, users are expected to understand device communication patterns and rewrite their workflows to be distribution-aware. In FastEstimator, all modules are designed to be distribution-friendly such that users can scale their training and evaluation across multiple devices without any change of code.  

\begin{figure}[ht] \centering \begin{subfigure}[t]{0.4308275\linewidth} \includegraphics[width=\linewidth]{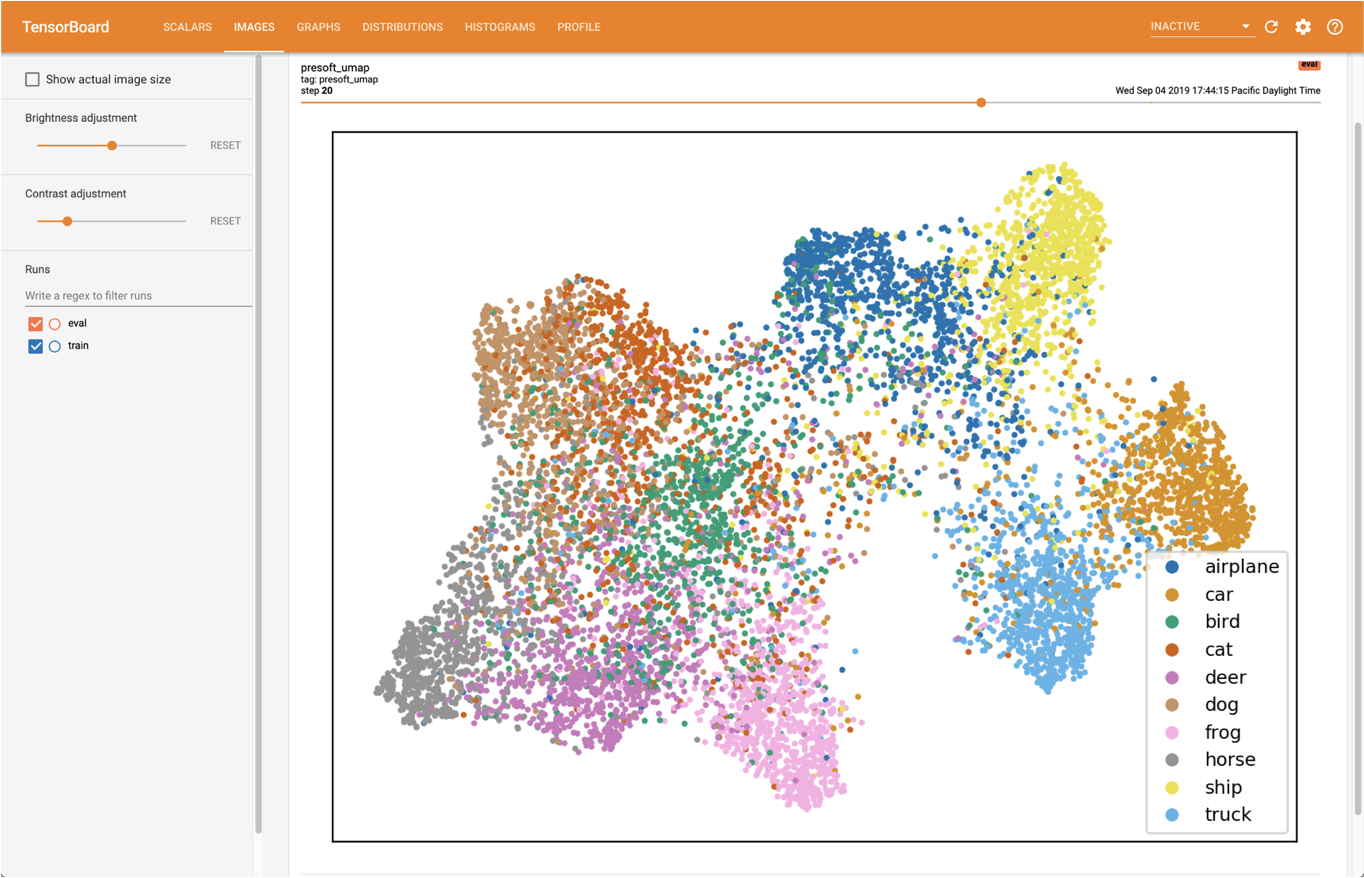} \caption{Activation UMAP}\label{fig:visA} \end{subfigure}%
\begin{subfigure}[t]{0.3048010\linewidth} \includegraphics[width=\linewidth]{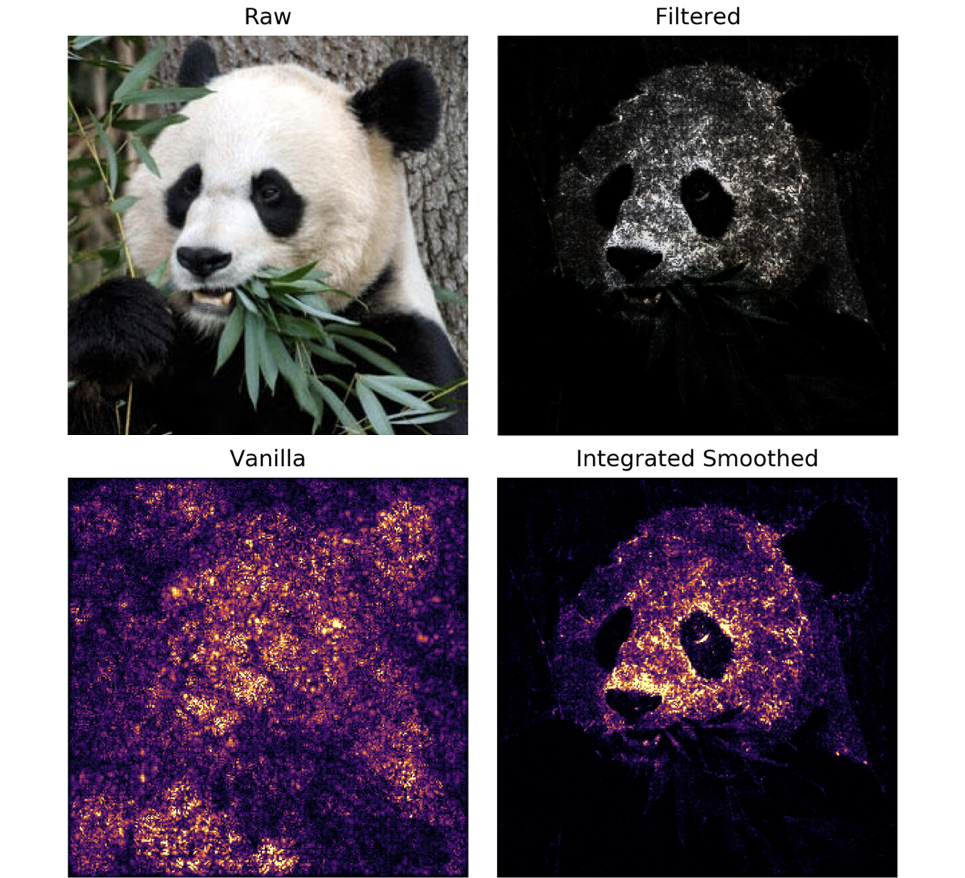} \caption{Saliency Map}\label{fig:visB} \end{subfigure}%
\begin{subfigure}[t]{0.2643714\linewidth} \includegraphics[width=\linewidth]{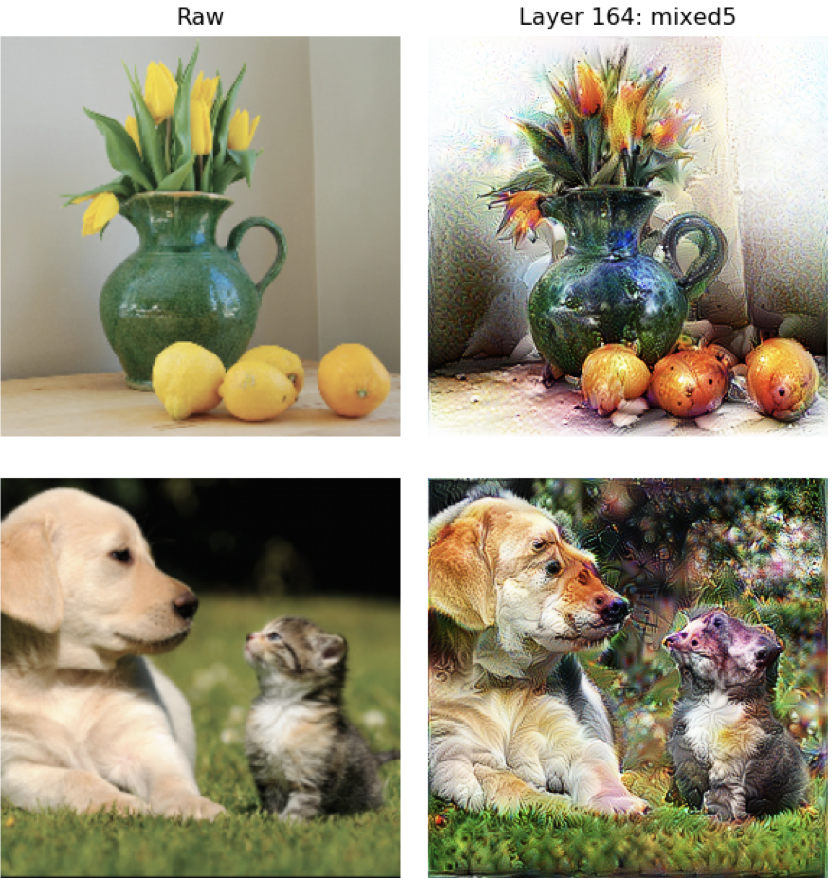} \caption{Caricatures}\label{fig:visC} \end{subfigure}
\caption{Visualization utility tools} \label{fig:vis} \end{figure}

\textbf{Useful Utility AI Tools.} Besides the training APIs, FastEstimator offers useful AI utility functions to facilitate the prototyping and production process. For example, the model interpretation module contains visualization tools such as feature UMAP~\cite{umap}, saliency maps~\cite{saliency1, saliency2} and caricature maps~\cite{feature-viz} to help users build more robust models (Fig.~\ref{fig:vis}). All utility modules have full compatibility with TensorBoard. We also provide utility tools for automatic report generation for documentation purposes.  

\textbf{Application Hub.} Some frameworks~\cite{tensorflow, mxnet, pytorch, keras} provide a model zoo, which allows users to import pre-built model architectures and weights. However, it is often the case that the true complexity of implementing a new idea lies more on the data pipeline and training loop than on the model architecture itself. For this reason,  FastEstimator provides Application Hub\textemdash a place to showcase different end-to-end AI applications. Every template in Application Hub has step-by-step instructions to ensure users can easily build new AI applications with their own data.   

\section{Architecture Overview} \label{sec:overview} 
\begin{figure}[ht] \centering \includegraphics[width=\linewidth]{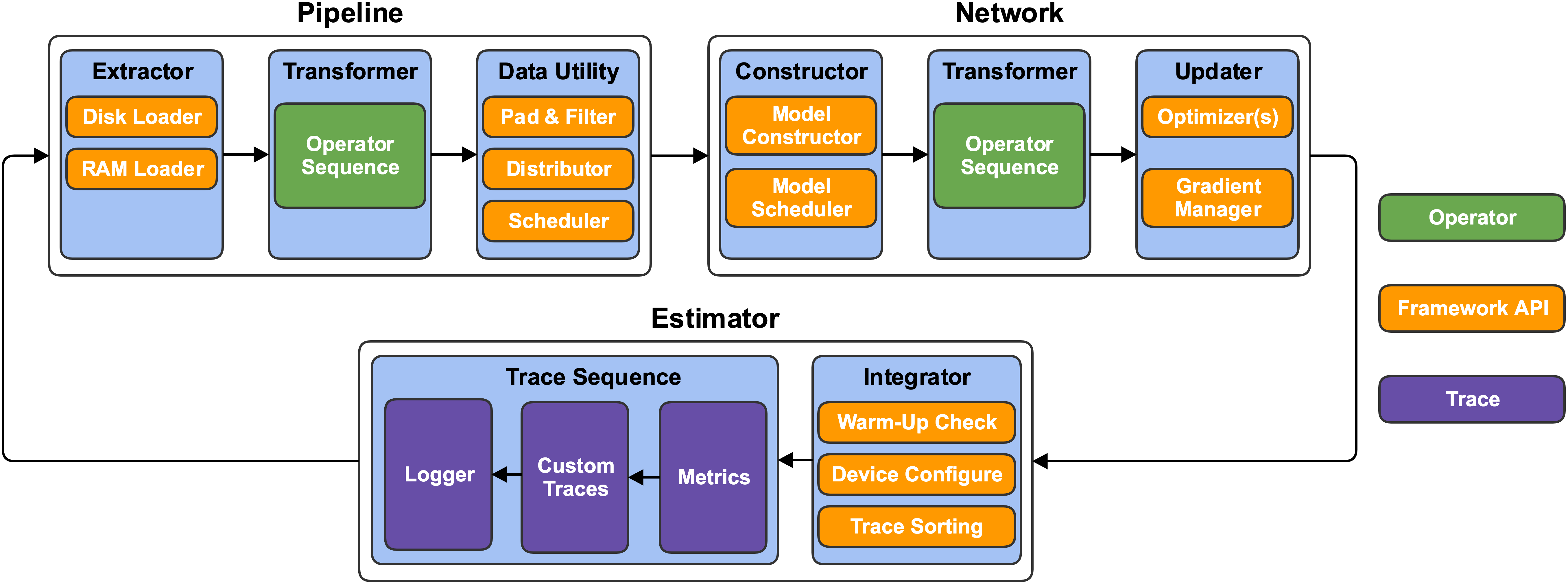} \caption{FastEstimator architecture overview} \label{fig:architecture}  
\end{figure}

All deep learning training workflows involve three essential components: data pipeline, network, and optimization strategy. Data pipeline extracts data from disk to RAM, performs transformations, and then loads the data onto the device. Network stores trainable and differentiable graphs. Optimization strategy combines data pipeline and network in an iterative process. Each of these components represents a critical API in FastEstimator: \texttt{Pipeline}, \texttt{Network}, and \texttt{Estimator} (Fig.~\ref{fig:architecture}). Users will interact with these three APIs for any deep learning task. 

\texttt{Pipeline} can be summarized as an Extraction-Transformation-Load (ETL) process. The extractor can take data either from disk or RAM, with features being either paired or unpaired (e.g., domain adaptation). The transformer builds graphs for preprocessing. The data utility provides support for scenarios like imbalanced training, feature padding, distributed training, and progressive training.  

\texttt{Network} manages trainable models. First, the constructor builds model graphs and creates timestamps on these graphs in the case of progressive training. The transformer then connects different pieces of model graphs and non-trainable graphs together. The updater tracks and applies gradients to each trainable model. 

\texttt{Estimator} is responsible for the training loop. Before training starts, a smoke test is performed on all graphs to detect potential run-time errors as well as to warm up the graph for faster execution. It then proceeds with training, generating any user-specified output along the way. 

The central component of both \texttt{Pipeline} and \texttt{Network} is a sequence of \texttt{Operators}. Similarly for the \texttt{Estimator}, there is a sequence of \texttt{Traces}. \texttt{Operator} and \texttt{Trace} are both core concepts which differentiate FastEstimator from other frameworks. 

\section{Core Concepts} \label{sec:core_concept}
\subsection{Operator} \label{subsec:operator}  
The common goal of all high-level deep learning APIs is to enable complex graph building with less code. For that, most frameworks like Keras~\cite{keras}, MXNet~\cite{mxnet}, and PyTorch~\cite{pytorch} introduce the concept of layers (aka blocks and modules) to simplify network definition. However, as model complexity increases, even layer representations may become undesirably verbose\textemdash for example when expressing multiple time-dependent model connections. Therefore, we propose the concept of \texttt{Operator}, a higher level abstraction for layers, to achieve code reduction without losing flexibility.  

\begin{figure}[ht] \centering \includegraphics[width=\linewidth]{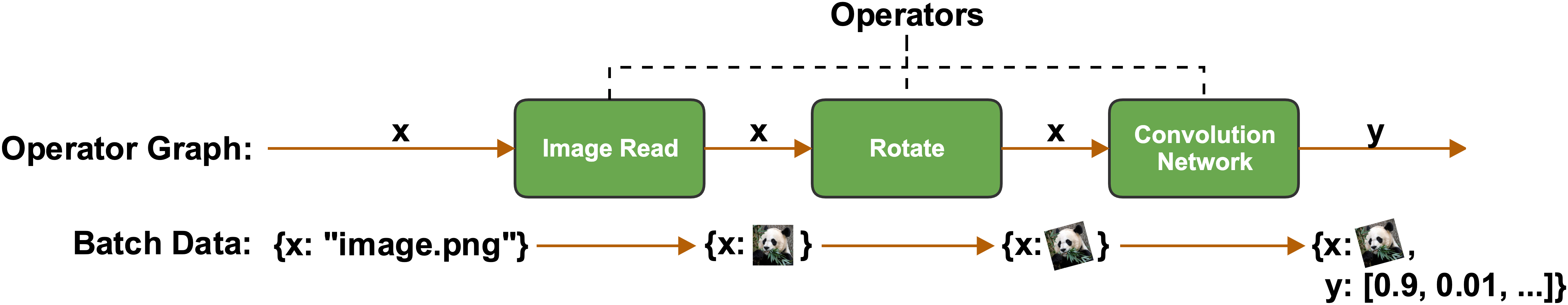} \caption{A sequence of \texttt{Operators} reading and modifying batch data} \label{fig:opChain} \end{figure}

An \texttt{Operator} represents a task-level computation module for data (in the form of key:value pairs), which can be either trainable or non-trainable. Every \texttt{Operator} has three components:  input key(s), transformation function, and output key(s).  The execution flow of a single \texttt{Operator} involves:  \begin {enumerate*}[1)]
\item take the value of the input key(s) from the batch data,
\item apply transformation functions to the input value,
\item write the output value back to the batch data with output key(s).  \end{enumerate*} Fig.~\ref{fig:opChain} shows how a deep learning application can be expressed as a sequence of \texttt{Operators}.  

\begin{figure}[ht] \centering \includegraphics[width=\linewidth]{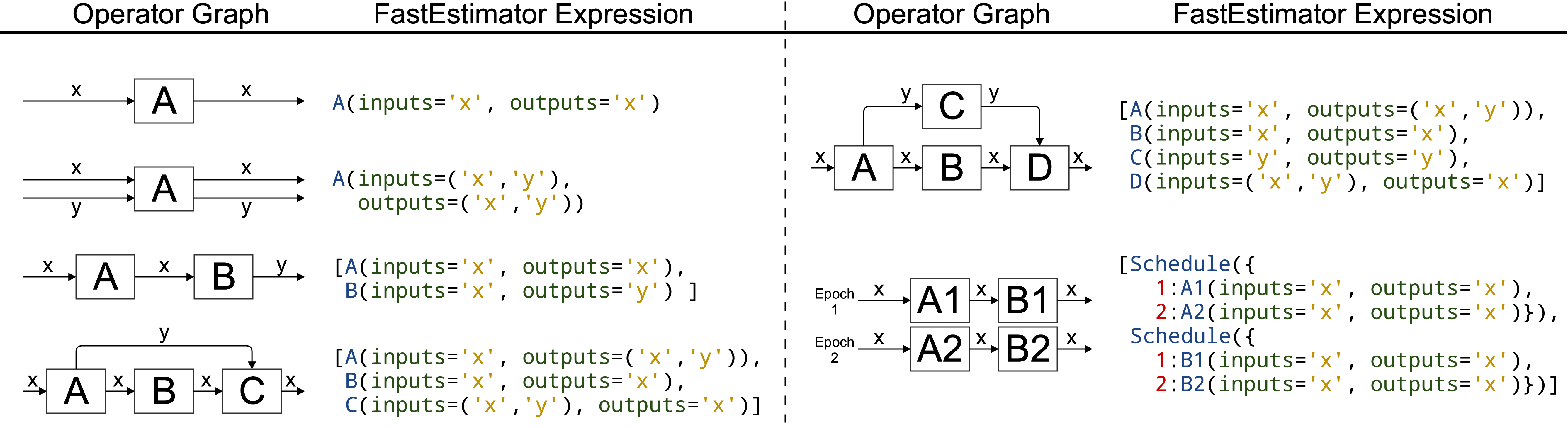} \caption{\texttt{Operator} graphs and FastEstimator expressions} \label{fig:ops} \end{figure}

FastEstimator offers concise \texttt{Operator} expressions that allow users to construct various graph topologies in an efficient way (Fig.~\ref{fig:ops}). With the help of \texttt{Operators}, complex computational graphs can be built using only a few lines of code. We will further demonstrate the convenience of \texttt{Operators} in Sec.~\ref{sec:examples} on different deep learning applications.  

\subsection{Trace} \label{subsec:trace}  
Some APIs~\cite{tensorflow,pytorch,keras} offer two separate modules: metrics and callbacks (aka hooks and handlers). In FastEstimator, both metrics and callbacks are unified into \texttt{Traces}. As a result, we can effectively overcome several limitations introduced by separating metrics and callbacks.   

Metrics are quantitative measures of model performance and are computed during the training or validation loop. From the implementation perspective, APIs tend to implement metrics as a built-in computational graph with two parts: a value and an update rule.  While this pre-compiled graph enables faster computation, it also limits the choice of metrics. For example, some domain-specific metrics are not easily expressed as a graph or require running external post-processing libraries. Furthermore, the benefit offered by pre-compiling metric graphs is not significant, because these calculations only account for a small portion of the system’s total computation.   

\begin{figure}[ht] \centering \begin{subfigure}[t]{0.433001531\linewidth} \includegraphics[width=\linewidth]{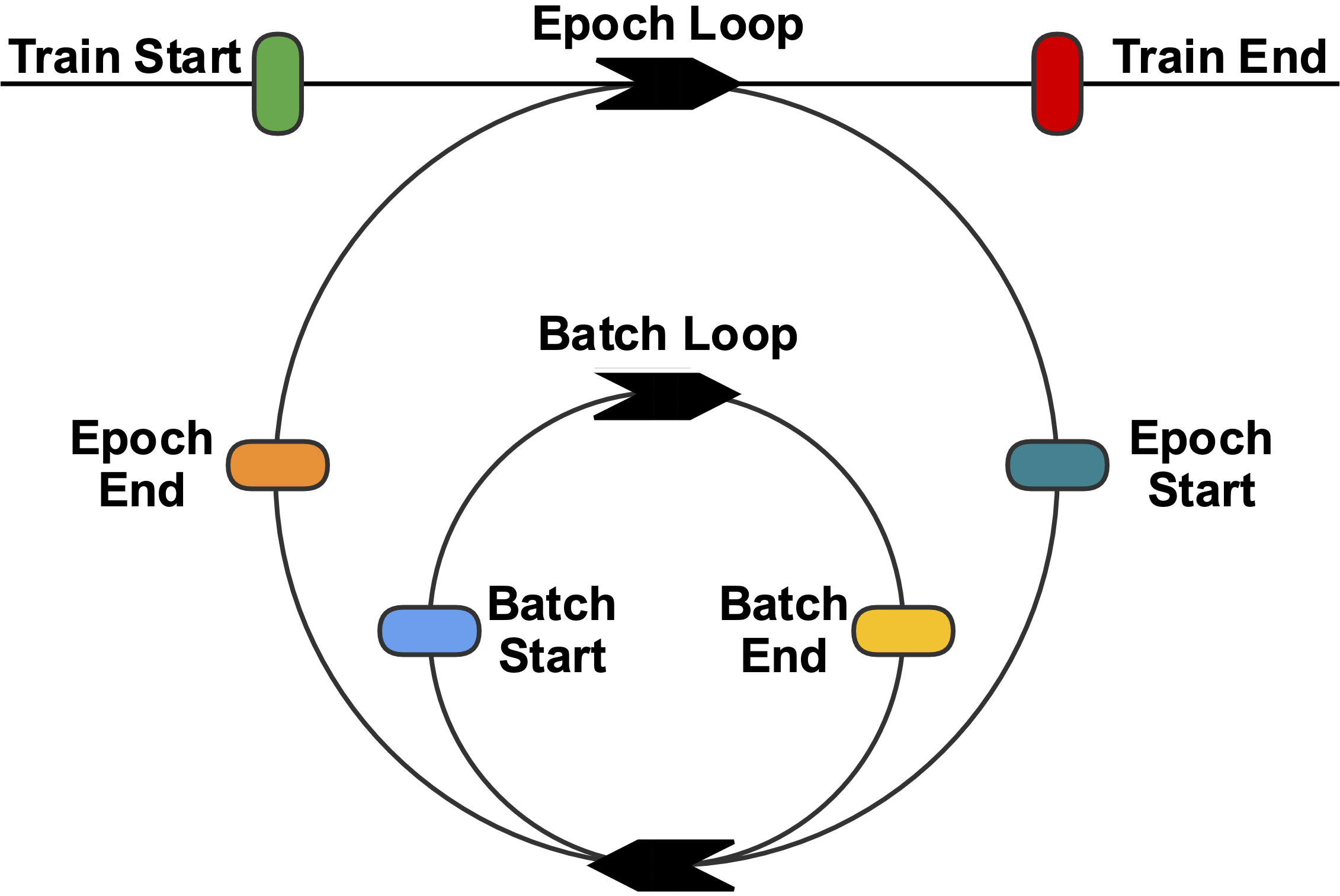} \caption{Events within the training loop}\label{fig:traceA} \end{subfigure}%
\begin{subfigure}[t]{0.187978560\linewidth} \includegraphics[width=\linewidth]{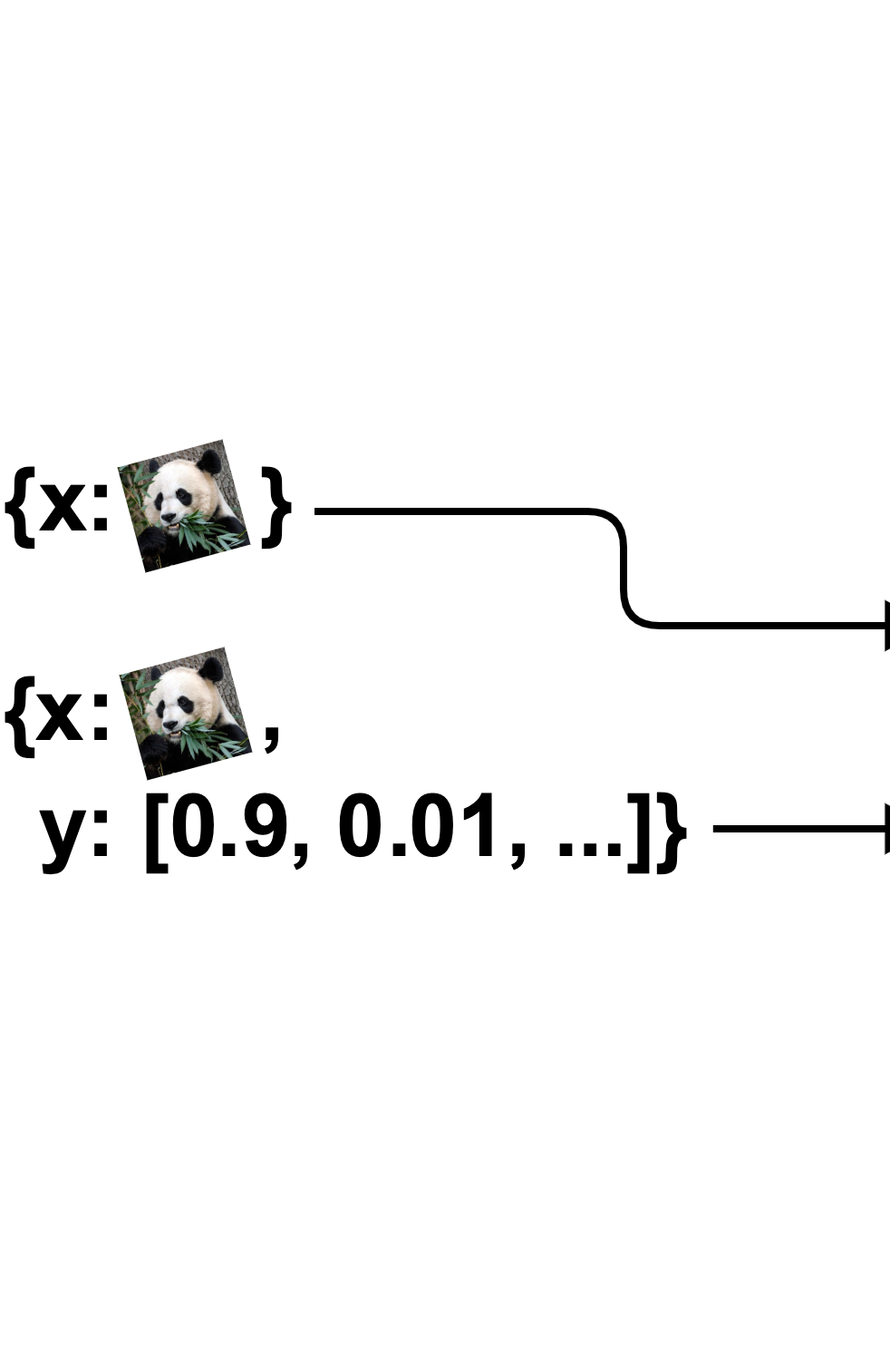} \end{subfigure}%
\begin{subfigure}[t]{0.379019908\linewidth} \includegraphics[width=\linewidth]{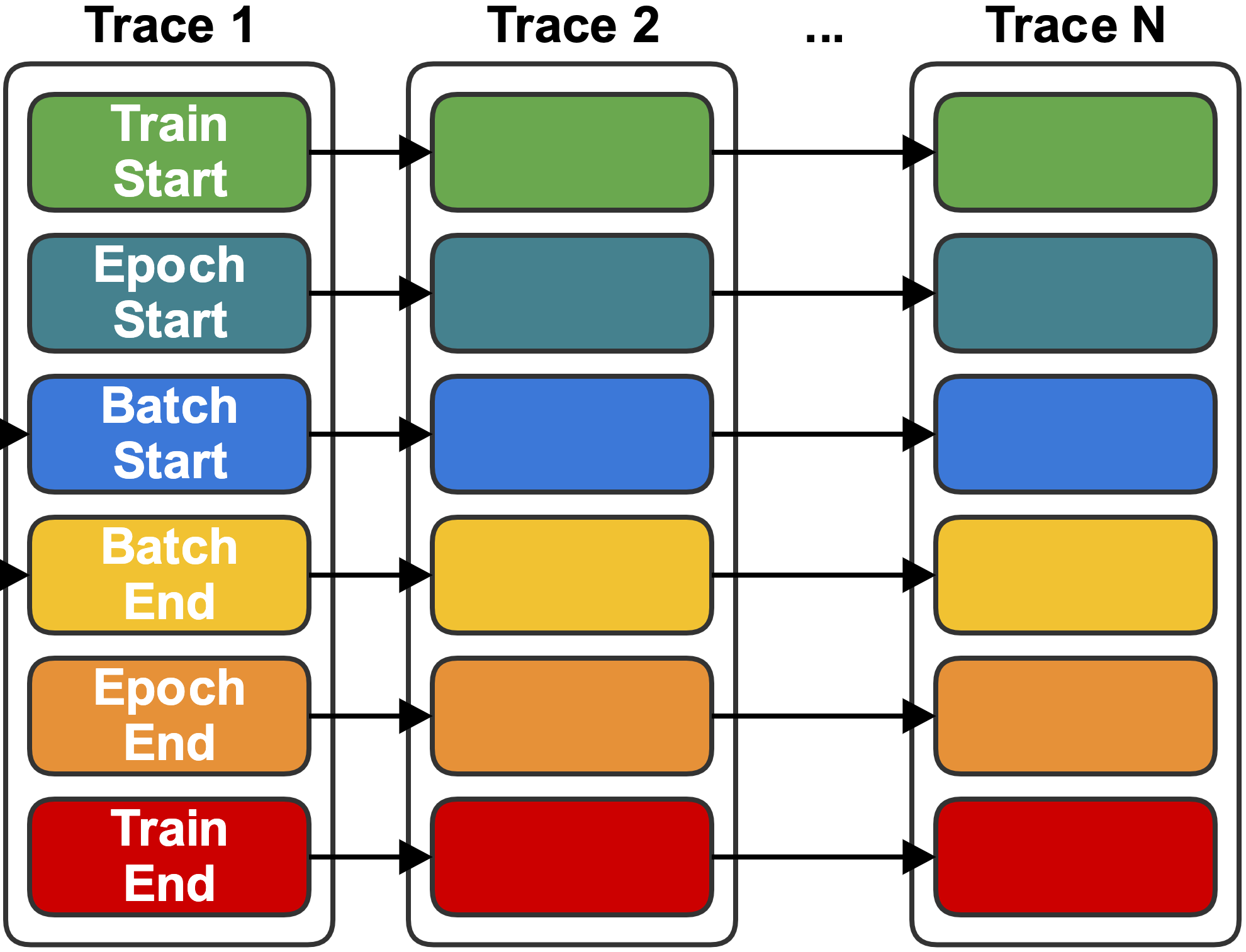} \caption{\texttt{Trace} communications on events}\label{fig:traceB} \end{subfigure} \caption{\texttt{Traces} throughout training} \label{fig:trace} \end{figure}

Callbacks are modules that contain event functions like \texttt{on\_epoch\_begin} and \texttt{on\_batch\_begin},  which allow users to insert custom functions to be executed at different locations within the training loop (Fig.~\ref{fig:traceA}).  Implementation-wise, since metrics and callbacks are separate, callbacks in most frameworks are not designed to have easy access to batch data. As a result, researchers may have to use less efficient workarounds to access intermediate results produced within the training loop. Moreover, callbacks are not designed to communicate with each other, which adds further inconvenience if a later callback needs the outputs from a previous callback.

\texttt{Trace} in FastEstimator is a unification of metrics and callbacks; it preserves the event functions in callbacks and overcomes the aforementioned limitations through the following improvements: \begin {enumerate*}[1)]
\item \texttt{Traces} have easy access to batch data directly from the batch loop,
\item every \texttt{Trace} can pass data to later \texttt{Traces} to increase re-usability of results as shown in Fig.~\ref{fig:traceB},
\item metric computation can leverage batch data directly without a graph, 
\item metrics can be accumulated through \texttt{Trace} member variables without update rules. \end{enumerate*} 
These improvements brought by \texttt{Trace} have enabled many new functionalities that are not easily achieved with conventional callbacks.  For example, our model interpretation module is made possible by easy batch data access. Furthermore, \texttt{Trace} has access to all API components such that changing model architecture or data pipeline within the training loop is straightforward. This can unlock support for Meta-Learning, Reinforcement Learning (RL), and AutoML algorithms. 

\section{Examples and Applications} \label{sec:examples} 
In this section, we show FastEstimator \texttt{Operator} expressions for various deep learning tasks. All source code is available on \href{https://github.com/fastestimator}{\texttt{github.com/fastestimator}}.

\begin{figure}[ht] \centering \includegraphics[width=0.85\linewidth]{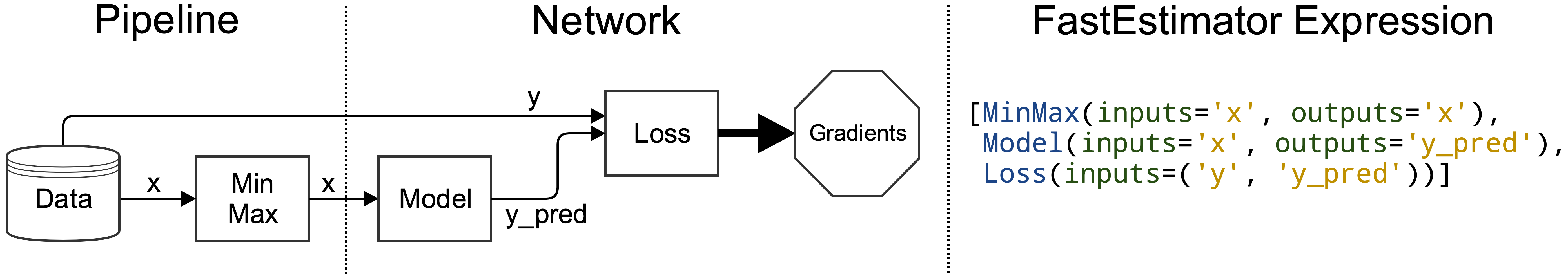} \caption{An image classification example, where the input image has key \texttt{x} and label key \texttt{y}} \label{fig:cls} \end{figure}

\textbf{Image Classification.} We start with an image classification example (Fig.~\ref{fig:cls}). \texttt{MinMax} applies normalization to input images before the model forward pass, the system calculates the gradients from the provided loss function, and performs back propagation. 

\begin{figure}[ht] \centering \includegraphics[width=0.85\linewidth]{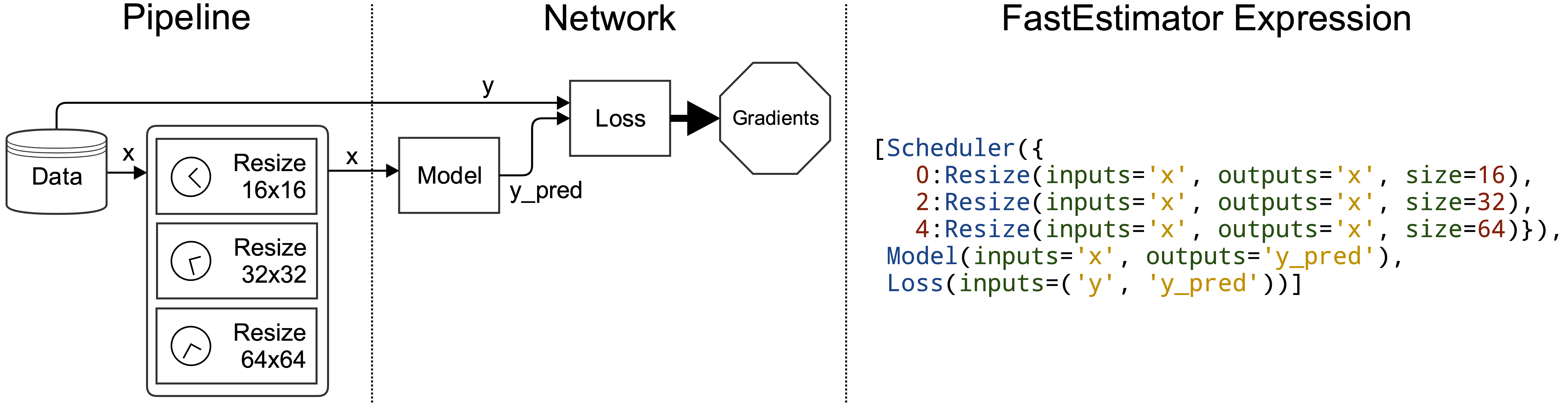} \caption{Image classification with progressive resizing} \label{fig:cls_pro} \end{figure}

\textbf{Image Classification with Progressive Resizing.} Progressive resizing techniques have been applied to super-resolution~\cite{super-res} and GAN training~\cite{pggan}. They start with low resolution images and gradually increase the image resolution as training progresses. Fig.~\ref{fig:cls_pro} shows an example where we increase the image resolution by a factor of 2 on epoch 2 and 4.

\begin{figure}[ht] \centering \includegraphics[width=\linewidth]{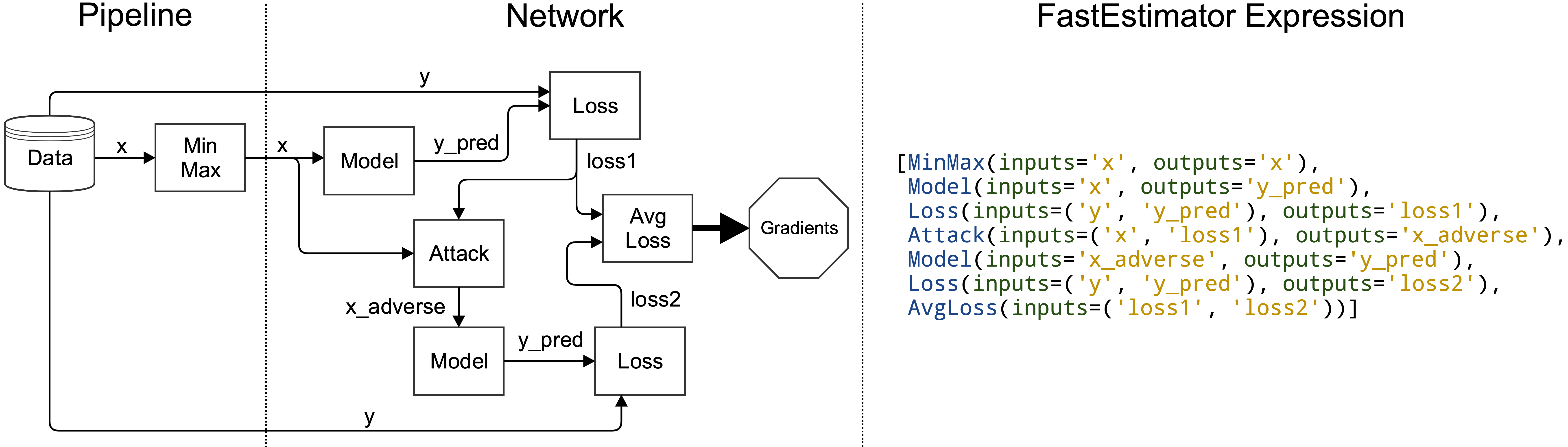} \caption{Image classification with adversarial training} \label{fig:cls_adv} \end{figure}

\textbf{Image Classification with Adversarial Training.} It is also easy to apply modifications to the training process during the network forward pass. For example, training a neural network using input images perturbed by an adversarial attack\textemdash which has been shown to make models more robust to future adversarial attacks~\cite{attack}\textemdash can be accomplished with only four extra \texttt{Operators}. This makes it easy to add adversarial training to any model (Fig.~\ref{fig:cls_adv}). 

\begin{figure}[ht] \centering \includegraphics[width=\linewidth]{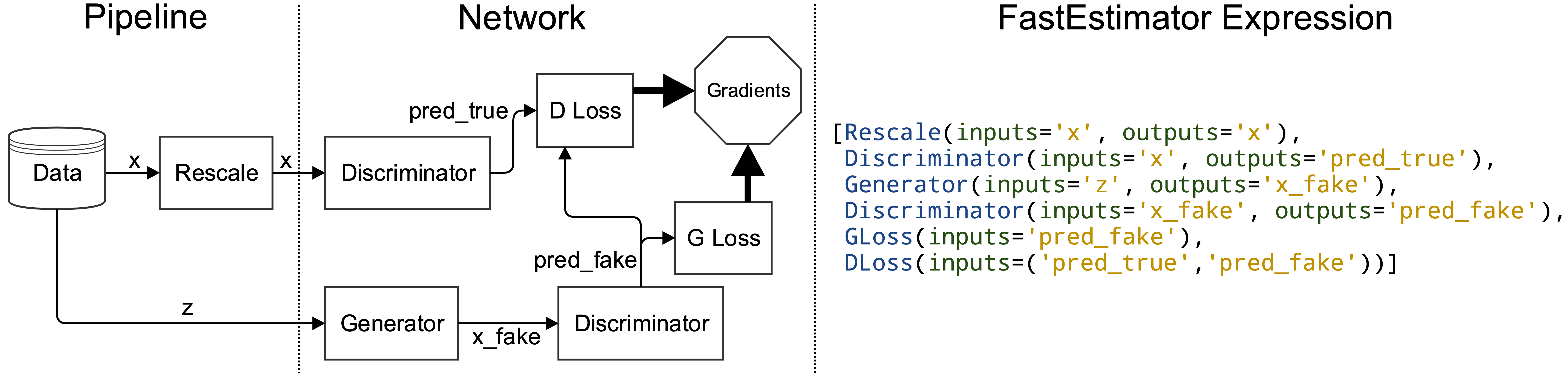} \caption{Image generation with DC-GAN, with real image as \texttt{x} and random noise as \texttt{z}} \label{fig:dcgan} \end{figure}

\textbf{Image Generation with Deep Convolutional GAN.} For multi-model training such as DC-GAN ~\cite{dcgan}, users can associate different losses to different models. The gradients are calculated with respect to each loss and the system will perform updates for each model (Fig.~\ref{fig:dcgan}).

\begin{figure}[ht] \centering \includegraphics[width=\linewidth]{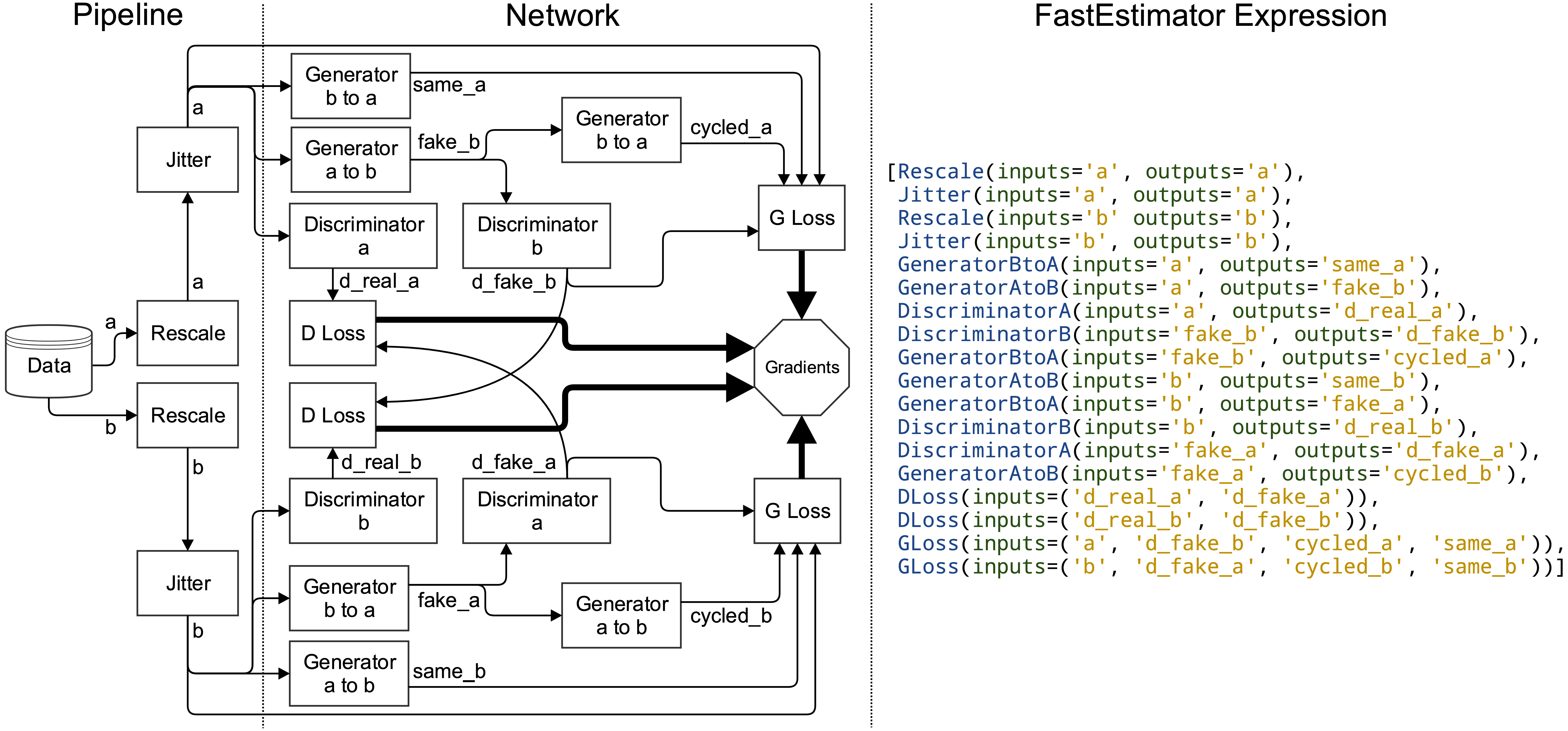} \caption{Image generation with Cycle-GAN, with \texttt{a} and \texttt{b} as unpaired images from two domains} \label{fig:cycleGAN} \end{figure}

\textbf{Image Generation with Cycle-GAN.} For a more complicated problem such as unsupervised unpaired image translation using the Cycle-GAN~\cite{cyclegan}, users often need to define different losses that involve outputs of multiple models.  FastEstimator \texttt{Operators} make it easy to break down such complex interactions between different generators and discriminators (Fig.~\ref{fig:cycleGAN}). 

\section{Future} \label{sec:future} 
Going forward we intend to expand FastEstimator into the deep RL domain by integrating our API with existing RL libraries such as those provided by OpenAI~\cite{openai} and TensorFlow~\cite{tensorflow}. We also intend to provide more support for AutoML and Meta-Learning. Finally, we will continue implementing and enabling more state-of-the-art ideas across different domains to provide users with ready-made AI solutions.   

\subsubsection*{Acknowledgments}

We would like to thank the following people for their feedback and guidance: Dibyajyoti Pati, Hans Krupakar, Min Zhang, Ravi Soni, Pál Tegzes, Levente Török, Dániel Szabó, Máté Fejes, Valentin Mikhaylenko, and Karley Yoder. 

\newpage

\bibliography{neurips_2019}{}  
\bibliographystyle{unsrt}  
\end{document}